\pdfoutput=1

\documentclass[11pt]{article}

\usepackage{todonotes}
\usepackage{multirow}
\usepackage{amsfonts}
\usepackage{amsmath}
\usepackage{breqn}
\usepackage{hhline}
\usepackage{booktabs}
\usepackage{url}

\usepackage[]{acl}
\usepackage{times}
\usepackage{latexsym}

\usepackage[T1]{fontenc}

\usepackage[utf8]{inputenc}

\usepackage{microtype}

\title{Dialogue Evaluation with Offline Reinforcement Learning}

\author{Nurul Lubis, Christian Geishauser, Hsien-Chin Lin, \\\textbf{ Carel van Niekerk, Michael Heck, Shutong Feng, Milica Ga\v{s}i\'{c}} \\
 Heinrich Heine University Düsseldorf, Germany \\
 \texttt{\{lubis, geishaus, linh, niekerk, heckmi, fengs, gasic\}@hhu.de}}

\begin{document}
\maketitle
\begin{abstract}
Task-oriented dialogue systems aim to fulfill user goals through natural language interactions. They are ideally evaluated with human users, which however is unattainable to do at every iteration of the development phase. Simulated users could be an alternative, however their development is nontrivial. Therefore, researchers resort to offline metrics on existing human-human corpora, which are more practical and easily reproducible. They are unfortunately limited in reflecting real performance of dialogue systems. BLEU for instance is poorly correlated with human judgment, and existing corpus-based metrics such as success rate overlook dialogue context mismatches. There is still a need for a reliable metric for task-oriented systems with good generalization and strong correlation with human judgements. In this paper, we propose the use of offline reinforcement learning for dialogue evaluation based on a static corpus.
Such an evaluator is typically called a critic and utilized for policy optimization. We go one step further and show that offline RL critics can be trained on a static corpus for any dialogue system as external evaluators, allowing dialogue performance comparisons across various types of systems.
This approach has the benefit of being corpus- and model-independent, while attaining strong correlation with human judgements, which we confirm via an interactive user trial.
\end{abstract}

\section{Introduction}
With the rise of personal assistants, task-oriented dialogue systems have received a surge in popularity and acceptance. Task-oriented dialogue systems are characterized by a user goal which motivates the interaction, e.g., booking a hotel, searching for a restaurant or calling a taxi. The dialogue agent is considered successful if it is able to fulfill the user goal by the end of the interaction.


Ideally, success rates are obtained via interaction with a real user in-the-wild. Unfortunately, with a handful of exceptions, e.g., LetsGO~\cite{lee2018dialcrowd} and Alexa Challenge~\cite{gabriel2020further}, that is often out of reach. The closest approximation is human trials with paid users such as Amazon Mechanical Turk workers, which has also been adopted as final evaluation in recent incarnations of the Dialogue State Tracking Challenge (DSTC) \cite{gunasekara2020overview}. However, such evaluations are highly time- and cost-intensive, making them impractical for optimization during an iterative development. The third alternative is to use a user simulator to conduct online dialogue simulation, however the result is subject to the quality of the user simulator itself.
Furthermore, developing such simulators is far from straightforward and requires significant amounts of handcrafting~\cite{schatzmann08a}. Only recently we have seen data-driven user simulators that can compete with hand-coded ones~\cite{lin-etal-2021-domain}.

While there has been considerable progress towards more meaningful automatic evaluation metrics for dialogues, there remains a number of limitations as highlighted by the recent NSF report \cite{mehri2022report}: the metrics 1) measure only a limited set of dialogue qualities, which mostly focus on subjective aspects such as fluency and coherence, 2) lack generalization across datasets and models, and 3) are not yet \textit{strongly} correlated with human judgements. These limitations hinder a more widespread use of newly proposed metrics for benchmarking and comparison, especially with prior works. Further, in particular for task-oriented dialogue systems, the need for reliable automatic evaluation of dialogue success is still unanswered.

Being able to automatically evaluate the success rate of any policy using static data offers a number of benefits in terms of required resources, generalizability, and reproducibility. Furthermore, it is not only suitable for the final evaluation of a dialogue policy, but can also be utilized as an objective for iterative optimization. The corpus-based success rate is one such method, which has become the standard metric for state-of-the-art comparisons of policy optimization approaches today~\cite{budzianowski2018large}. Unfortunately, this metric is computed based on pseudo-dialogues that may contain context mismatch. Therefore, we believe it should be treated more as an approximation: it is insufficient at best, and misleading at worst, in reflecting real performance of dialogue systems. In addition, the rules used to check the goal completion need to be handcrafted based on the ontology, making this method data- or ontology-dependent.

In this paper, we propose to use offline reinforcement learning (RL) to train a policy evaluator, also known as a critic, based on a static collection of dialogue data\footnote{ https://gitlab.cs.uni-duesseldorf.de/general/dsml/lava-plas-public}. We show that an offline critic addresses the limitations of current automatic metrics: 1) it can be trained to evaluate any dialogue system architecture after-the-fact, allowing comparisons across various types of systems from prior works, 2) it can be utilized in the iterative development phase to optimize a dialogue policy, 3) it is theoretically grounded, 
solving the problems that standard corpus-based success rate has due to context mismatch, and 4) it strongly correlates with the performance of the system when interacting with human users, which we confirm via a user trial.

\section{Related Work}

For a long time, the research in dialogue policy has focused on user-centered criteria such as user satisfaction~\cite{walker1997paradise,lees12,ultes2017domain}. The most reliable way to obtain these scores is to have users interact directly with the system and let them subjectively rate the system afterwards. Due to the time and resource requirements to carry out such evaluations, human trials are usually done only as the final evaluation after the system development is finished.

As the line between policy and natural language generation (NLG) tasks becomes blurred, we see the introduction of metrics such as BLEU~\cite{papineni2002bleu} and perplexity. However, these have been labeled early on to be potentially misleading, as they correlate poorly with human judgement~\cite{stent2005evaluating,liu2016not}. This circumstance motivates automatic metrics that are highly correlated with human ratings~\cite{dziri-etal-2019-evaluating,mehri2020unsupervised,mehri2020usr}. However, these metrics are designed to measure subjective quality of a dialogue response, making them more suitable for evaluating chat-based systems.

Despite the availability of toolkits that facilitate user simulation (US) evaluation~\cite{zhu2020convlab}, corpus-based match and success rates are the default benchmark for works in task-oriented dialogue systems today~\cite{budzianowski2018large,nekvinda2021shades}. These metrics are practical to compute, reproducible, and scalable. 
Current standard corpus-based metrics are computed on a pseudo-dialogue constructed using user utterances from data and responses generated by the system. A set of rules then checks whether the system provides all information requested by the user. Unfortunately, they do not take into account context mismatches that may originate from the pseudo-dialogue construction and therefore does not reflect other aspects of dialogue quality as the resulting dialogue flow is completely overlooked. 

There has been few applications of offline RL to dialogue systems. \citet{jaques2019way} explores various language-based criteria, e.g., sentiment and semantic similarity, as reward signals for open-domain dialogue, paired with a Kullback-Leibler (KL) control for exploration within the support of the data. \citet{verma2022chai} proposed using fine-tuned language models to utilize unlabeled data for learning the critic function. The method is however only demonstrated on a very small state and action space, and it is therefore unclear whether it generalizes to more complex set ups. \citet{ramachandran2021causal} applied offline RL with a pair-wise reward learning model based on preference learning, however it still utilizes the corpus-based success rate for choosing the preferred rollout. 
To the best of our knowledge, offline RL has not previously been deployed for dialogue evaluation.

\section{Preliminaries}
\label{sec:offlineRL}

\subsection{Offline RL}
Dialogue can be formulated as a reinforcement learning problem with a Markov decision process (MDP) $\mathcal{M}=\{\mathcal{S}, \mathcal{A}, r, p, p_0, \gamma\}$. In this MDP, $\mathcal{S}$, $\mathcal{A}$, and $r$ denote the state and action spaces, and the reward function, respectively. $p(s_{t+1}|s_t, a_t)$ denotes the probability of transitioning to state $s_{t+1}$ from $s_t$ after executing $a_t$, and $p_0(s)$ is the probability of starting in state $s$. $\gamma \in [0, 1]$ is the discount factor that weighs the importance of immediate and future rewards. At each time step $t$, the agent observes a state $s_t$, executes its policy $\pi$ by selecting an action $a_t$ according to $\pi(a_t|s_t)$, transitions to a new state $s_{t+1}$ and receives a reward $r_t$. 
The goal of the policy is to maximize the cumulative discounted rewards, i.e., the return $R_t=\sum_{i \geq 0}\gamma^ir_{t+i}$.

Instead of interacting with the MDP to learn a policy, offline RL aims to learn a policy exclusively from previously collected data containing state transitions $\mathcal{D} = \{(s_i, a_i, s_{i+1}, r_i)\}_i$ under an unknown behavior policy $\pi_\beta$. This set-up is especially useful in cases where deploying the agent in the real environment is too costly, as is the case with real user interaction for dialogue systems. As the agent can not interact with the environment, the performance of the trained policy $\pi$ needs to be evaluated also based on the data $\mathcal{D}$.
The Q-value $Q_\pi(s_t, a_t)$ denotes the expected return when executing $a_t$ in $s_t$ and following policy $\pi$ thereafter. 
Q-learning algorithms estimate the Q-function $Q_\pi$ by iteratively applying the Bellman operator
\begin{equation}
    \mathcal{T} Q(s_t, a_t) = \mathbb{E}_{s_{t+1}} [r_t + \gamma Q(s_{t+1}, a_{t+1})].
\end{equation}
Value-based RL methods optimize the policy by maximizing the Q-values for every state-action pair $(s_t, a_t) \in \mathcal{S} \times \mathcal{A}$. With discrete actions, and for given state $s$, the actor can then simply select $\textup{argmax}_{a} Q(s, a)$ in a greedy fashion. 


Alternatively, with an actor-critic method, an actor is trained which optimizes its parameters to maximize the expected return of the starting states, for example via the deterministic policy gradient method \cite{silver2014deterministic, lillicrap2016continuous}:
\begin{equation}
	\nabla_\theta J(\theta)=\mathbb{E}_{s \sim \mathcal{S}}[\nabla_\theta \pi_\theta(s) \nabla_a Q_\pi(s, a)|_{a=\pi(s)}].
	\label{eq:dpg}
\end{equation}

The challenge in performing offline RL comes from the fact that $\mathcal{D}$ is static and has limited coverage of $\mathcal{S}$ and $\mathcal{A}$. While an out-of-distribution state is not a problem during training as the state is always sampled from $\mathcal{D}$, the policy may select an out-of-distribution action that is not contained in $\mathcal{D}$. This tends to lead to arbitrarily high estimates which further encourages the policy to take out-of-distribution actions. 
There are two main methods to counteract this: 1) constraining the policy to stay within the support of the dataset \cite{wu2019behavior,jaques2019way,fujimoto2019off,PLAS_corl2020}, and 2) modifying the critic to better handle out-of-distribution actions \cite{kumar2019stabilizing,kumar2020conservative}. In this work, we focus on the former. 


\subsection{Dialogue Policy in the Latent Action Space}
RL can be applied to a dialogue system policy at different levels of abstraction. Semantic actions, i.e., tuples containing intent, slot and values, such as \texttt{inform(area=centre)}, are widely used for a compact and well-defined action space \cite{GeishauserHLLHF21,tseng2021transferable}. Pre-defining the actions and labeling the dialogue data however requires considerable labor. In addition, the final policy needs to be evaluated dependent on an NLG module. On the opposite end, natural language actions view each word of the entire system vocabulary as an action in a sequential decision making process \cite{mehri2019structured,jaques2019way}. This blows up the action space size and the trajectory length, hindering effective learning and optimal convergence.

\citet{zhao2019rethinking} proposed instead an automatically inferred latent space to serve as action space of the dialogue policy, where a latent action is a real-valued vector containing latent meaning. This decouples action selection and language generation, as well as shorten the dialogue trajectory.
\citet{lubis2020lava} followed up this work by proposing the use of variational auto-encoding (VAE) for a latent-space that is action characterized. 
In both of these works, the latent space is trained via supervised learning (SL) on the response generation task, and then followed with policy gradient RL using the corpus-based success as the reward signal, i.e.,
%
\begin{equation}
	\nabla_\theta J(\theta)=\mathbb{E}_\theta[\sum_{t=0}^{T}R_t\nabla_\theta\log p_\theta(z_t|c_t)].
	\label{eq:reinforce}
\end{equation}
%

%
\subsection{Offline RL for Policy in the Latent Action Space (PLAS)}
A latent action space also lends itself well to offline RL with a policy-constraint technique. \citet{PLAS_corl2020} proposed to use a conditional VAE (CVAE) to model the behavior policy $\pi_\beta(a|s)$ to reconstruct actions conditioned on states. 
The benefit of learning in the latent space is that the latent policy has the flexibility of choosing the shape of the distribution via the prior.
By constraining the latent policy to output latent actions with high probability under the prior, the decoder will output an action that is likely under the behavior policy in expectation. By choosing a simple prior such as a normal Gaussian distribution, constraint to the latent policy becomes simple to enforce, for example by defining $z = \pi(s)$ such that $z_i \in [-\sigma, \sigma]$ for each dimension $i$ of the latent space for some hyperparameter $\sigma$.
\begin{figure*}
	\centering
	\includegraphics[trim=0.0cm 0.1cm 0.0cm 0.05cm, clip=true, width=0.9\linewidth]{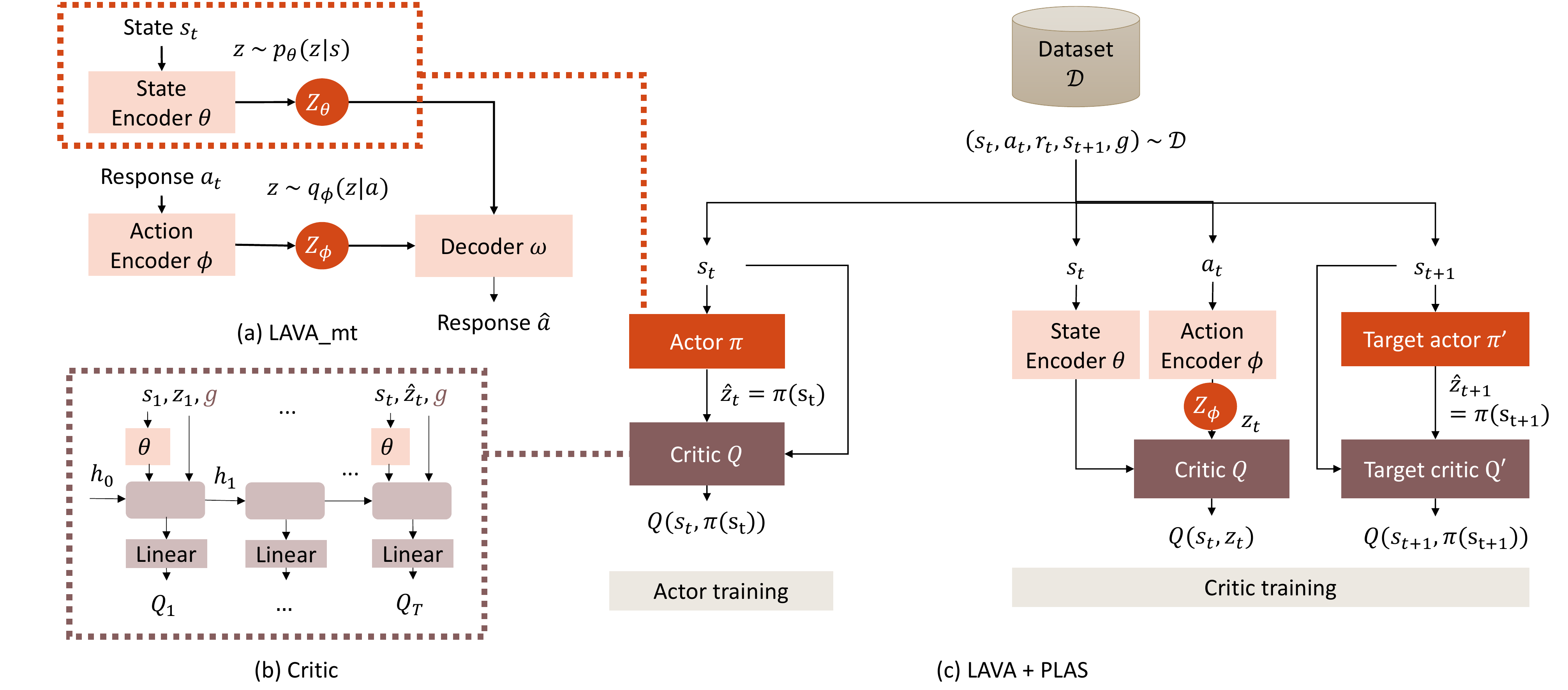}
	\caption{Overview of LAVA\_mt, critic network and offline RL with PLAS. First, (a) we pre-train LAVA\_mt with modified shared objective. The state encoder and latent space of the resulting model is used to initialize the actor for PLAS. The critic (b) is an RNN-based model that takes state, action and user goal to estimate the return. PLAS samples the transition from the static dataset and uses it to train actor and critic in an alternating fashion. To compute the target Q-value $Q(s_{t+1}, \pi(s_{t+1}))$, target actor and critic networks are used with soft update to improve stability.}
	\label{fig:overview}
	\vspace{-11pt}
\end{figure*}
PLAS defines a deterministic policy with continuous latent action that is optimized using the deterministic policy gradient method \cite{silver2014deterministic}. Dual critics are used that are optimized with soft clipped double Q-learning. The PLAS algorithm has been applied to real robot experiment as well as locomotive simulations tasks. In this environment, the latent actions and action space are continuous. This differs quite considerably from dialogue systems, where the latent action needs to be translated to word-level actions which are discrete.


\section{Offline Critic for Dialogue Policy Evaluation and Optimization}

The architecture of our proposed critic is depicted in Figure \ref{fig:overview}(b). We utilize recurrency to let the critic take dialogue context into account. 
We encode the word-level user utterance with an RNN and concatenate it with the binary belief state to obtain $s_t$.
On the other hand, the critic has the flexibility of taking any form of action. With latent actions, the action can be used as input directly by concatenating it with the state. When word-level or semantic actions are considered, a separate encoder can be used before concatenating it with the state.

In addition, to leverage the available data as much as possible, we incorporate the user goal for estimating the return. The MDP then becomes the dynamic parameter MDP (DP-MDP) as described by \citet{xie2020deep}, where a set of task parameters $g \in \mathcal{G}$ governs the state dynamics $p(s_{t+1}|s_t, a_t; g)$ and reward function $r(s_t, a_t;g)$. It is safe to incorporate the user goal for learning, because the critic is only used for policy evaluation and not needed to run the policy. If the user goal is not given in the data, it can be automatically derived from the dialogue state.
%
To maintain the correctness of the dialogue context, when predicting $Q(s_t, a_t)$, all actions $a_{< t}$ are taken from the corpus. Only $a_t$ is taken from the output of the policy. This is in contrast to the existing corpus-based success rate computation, where all $a_{\leq t}$ are taken from the policy and thus create context mismatches.

To keep the critic pessimistic in the face of uncertainty, we implement a dropout layer and do $K$ forward passes for each state-action pair and the lowest value is then taken as the final prediction, i.e., $Q(s_t, a_t) = \min_{k=1}^K Q_k$. In this way, prediction with high variance, i.e., high uncertainty, is punished by taking the lower bound. This mechanism replaces the use of double critic in PLAS.

\subsection{Offline Critic for Optimization: LAVA+PLAS}
\label{ssec:plaslava}
We combine LAVA \cite{lubis2020lava} and PLAS \cite{PLAS_corl2020} approaches in order to train a dialogue policy with latent action via offline RL. We use the multi-task LAVA approach, i.e., LAVA\_mt, depicted in Figure \ref{fig:overview}(a), using continuous latent variables modeled via Gaussian distributions, as the normal distribution prior works best with the PLAS approach. In the original LAVA\_mt, the model utilizes response generation (RG) and response VAE objectives for optimization with a 10:1 ratio, i.e., the VAE objective is optimized once every 10th RG epoch. In other words, the VAE is only used as an auxiliary task to ground the latent space from time to time. In this work, we modify the model training to preserve both RG and VAE abilities equally, as we will need the VAE to retrieve the latent action from the dataset $\mathcal{D}$. 

With $\theta$ as state encoder parameters, $\phi$ action encoder, and $\omega$ decoder, for each training pass, both tasks are performed and the model uses their joint loss to update its parameters, i.e.,
\begin{equation}
\begin{split}
  &\mathcal{L}_{\textup{LAVA\_mt}}(\omega, \theta, \phi) = \\
  &  \mathbb{E}_{p_{\theta}(z|s)}[\log p_\omega(x|z)]  -
   \alpha \textup{D}_{\textup{KL}}[p_{\theta}(z|s)||p(z)] \\& + \mathbb{E}_{q_{\phi}(z|r)}[\log p_\omega(x|z)]   - \beta \textup{D}_{\textup{KL}}[q_{\phi}(z|r)||p(z)].
\end{split}
\end{equation}
%
%
%
While the original LAVA uses policy gradient RL with the corpus-based success rate, in this work we follow the SL with PLAS algorithm. Parts of the LAVA\_mt model are used to initialize the actor and critic networks: parameters $\theta$ are used for the actor, $\phi$ to retrieve the latent action $z$ given a word-level response $a$, and the decoder $\omega$ to map latent actions produced by the actor into word-level responses. Prior to PLAS training, we warm-up the LAVA\_mt model with only the VAE objective to further improve the latent action reconstruction capability:
\begin{align}
\begin{split}
  & \mathcal{L}_{\textup{LAVA\_mt}}^{\textup{VAE}}(\omega, \phi) =
  \mathbb{E}_{q_{\phi}(z|r)}[\log p_\omega(x|z)] - \\& \beta \textup{D}_{\textup{KL}}[q_{\phi}(z|r)||p(z)].
  \label{eq:lava_mt_ae}
\end{split}
\end{align}

PLAS training is depicted in Figure \ref{fig:overview}(c). It consists of two interleaved training loops. For each pass, an episode is sampled from the static dataset $\mathcal{D}$. In the actor training loop, the actor parameter is optimized using deterministic policy gradient \cite{silver2014deterministic} to maximize the critic estimate. Due to the deterministic nature of the policy, the actor no longer samples from the distribution, but instead takes the distribution mean as the action. To encourage the policy to stay close to the behavior policy, as an additional loss, we add a mean-squared error (MSE) term between the chosen action $\hat{z}_t = \pi(s)$ and the reconstructed action from the corpus $z_t$. The actor loss is defined as
\vspace{-3pt}
\begin{equation}
	\mathcal{L}_{\textup{actor}} = Q(s, \pi(s)) + \textup{MSE}(\hat{z}_{t}, z_{t}).
	\label{eq:actor_loss}
\end{equation}
On the other hand, the critic is trained to minimize the error of the Bellman equation. In addition, we penalize the critic with a weighted KL loss term as a means of regularization when the target actor chooses an action that is far from the behavior policy. The critic loss is defined as
%
\begin{multline}
    \mathcal{L}_{\textup{critic}} = (Q(s_t, a_t) - (r_t + \gamma Q'(s_{t+1}, \pi'(s_{t+1})))^2 \\ - \lambda D_{KL}(q_\phi(z_{t+1}|a_{t+1})||\pi'(s_{t+1})).
    \label{eq:critic_loss}
\end{multline}
As is common practice, we use the target critic and actor networks for computing the target Q-value. The actor, critic, and their corresponding target networks are initialized the same way, but the target networks are updated with a soft update to promote stability in training.

\subsection{Offline Critic for Evaluation}\label{ssec:offline_eval}

In this paper, we utilize offline RL critic in a new way, as a data- and model-independent evaluator for task-oriented dialogue systems. Following the critic training loop in Figure \ref{fig:overview}(c), we replace the target actor with the fixed policy $\pi_e$, i.e. the one to be evaluated, and perform the critic loop training with Equation \ref{eq:critic_loss} as the loss function, setting $\lambda = 0$ for systems with word-level action. 

Note that with this approach, the dataset consisting of $N$ dialogues $\mathcal{D} = \{\{(s_i, a_i, s_{i+1}, r_i)\}_{i=1}^{T^n}\}_{n=1}^N$ for evaluation can take any form as long as the states $s_i$ and actions $a_i$ are compatible with the dialogue system input and output, allowing comparisons across various types of dialogues systems. For instance, the states $s_i$ can be represented as sequences of utterances or binary vectors and actions $a_i$ as word-level, latent, semantic, or binary actions. In terms of rewards, those can be sparse (i.e. intermediate rewards are set to $0$, $r_i=0$, $i<T_n$, $n=1,\dots,N$) and in case that the corpus represents the desireable behaviour, a maximum reward can be assumed as a final reward for every dialogue in the corpus (i.e. set to $1$, $r_{T_n}=1$, $n=1,\dots,N$). Of course more accurate reward labels would result in an even more precise evaluator. As a consequence, dialogue systems can be evaluated on static corpora that differ from the training corpus and also not necessarily generated by interacting with the system.  

A possible use case scenario would be a human-human corpus annotated with states and sparse rewards and a number of different dialogue systems being evaluated on this corpus. This is the case we consider in our evaluation below, whereby we use word-level and latent actions, and thus do not require explicit action labels.

\section{Experimental Set-up}
\subsection{Data}
We use MultiWOZ 2.1~\cite{budzianowski2018large, eric2019multiwoz} to conduct our experiments, one of the most challenging and largest corpora of its kind. MultiWOZ is a collection of conversations between humans in a Wizard-of-Oz fashion, where one person plays the role of a dialogue system and the other one a user. The user is tasked to find entities, e.g., a restaurant or a hotel, that fit certain criteria by interacting with the dialogue system. The corpus simulates a multi-domain task-oriented dialogue system interaction.%
We use the training, validation and test set partitions provided in the corpus, amounting to $8438$ dialogues for training and $1000$ each for validation and testing. 

\subsection{Policy and Critic Training} 
For the LAVA\_mt pre-training, we use simple recurrent models as encoder and decoder and follow the hyperparameters as set in the original work \cite{lubis2020lava} with a few exceptions, i.e.
we use $200$-dimensional continuous latent variables with a normal Gaussian as the prior and we lower the learning rate to $5\mathrm{e}{-4}$. As depicted in Figure \ref{fig:overview}, parts of the LAVA\_mt model are then used by the actor, critic, and different parts of PLAS training. For the critic, we set the hidden size to be $500$ and the linear layer to use the sigmoid activation function. During PLAS, we use a learning rate of $0.01$ for the critic and $0.005$ for the actor. The critic dropout rate and $\lambda$ are set to $0.3$ and $0.1$, respectively. The policy is trained with a maximum of 10K sampled episodes from the corpus, and the best checkpoint is chosen according to the corpus-based success rate. We set the hyper-parameters of the critic as an offline evaluator the same way, except that it uses 100K sampled episodes for training without early stopping.

\subsection{Dialogue Systems}

To show the generalization ability of our proposed offline evaluation, we evaluate various dialogue systems that differ in terms of modular abstractions and architectures:

\paragraph{HDSA \cite{chen2019semantically}}
is a transformer-based dialogue generation architecture with graph-based dialogue action using hierarchically-disentangled self-attention (HDSA). The model consists of a predictor, which outputs the dialogue action, and a generator, which subsequently maps it into dialogue response. Two versions of HDSA are included, one which uses ground-truth action for generation (gold), and one which uses predicted labels (pred). Note that the `pred' version is the only one that can be deployed in an interactive set-up.

\paragraph{AuGPT \cite{kulhanek2021augpt}}
is a fully end-to-end dialogue system with fine-tuned GPT2~\cite{radford2019language} on multi-task objectives, including belief state prediction, response prediction, belief-response consistency, user intent prediction, and system action prediction. The model is trained on MultiWOZ data augmented with the Taskmaster-1 \cite{byrne2019taskmaster} and Schema-Guided Dialogue \cite{rastogi2020towards} datasets.

\paragraph{LAVA \cite{lubis2020lava}}
is an RNN-based model using latent actions, optimized via SL and policy gradient RL with corpus-based success rate as reward. We use LAVA\_kl as the best performing model reported.

\paragraph{LAVA+PLAS (Ours)}
is our proposed variant of LAVA that is trained in an offline RL set-up using offline critic and PLAS algorithm (Section \ref{ssec:plaslava}).

\subsection{Evaluation Metrics}

\paragraph{Offline Critic for Evaluation (Ours)} 
For each system, we train an offline critic using offline Q-learning as described in Section \ref{ssec:offline_eval}. While theoretically the critic can take any form of dialogue action as input, in our experiments we utilize word-level or latent action. We consider intermediate rewards to be $0$ and the final reward is $1$ for a successful dialogue or $0$ for a failed dialogue, as provided in the MultiWOZ corpus.
As final estimated value of the policy, we report the average estimated return of all initial states on the test set.

\paragraph{Standard corpus-based metrics}

Corpus based evaluation is conducted on MultiWoZ test set using delexicalized responses with the benchmarking evaluation script provided by \citet{budzianowski2018large}. A pseudo dialogue is generated, where user turns are taken from the corpus and system turns are generated by the evaluated model. Match rate computes whether all informable slots in the user goal are generated, and success rate computes whether all information requested by the user is provided. For completeness, we also report the BLEU score on target responses.

\paragraph{US evaluation}

We use the default ConvLab2~\cite{zhu2020convlab} user simulator with the BERT-based NLU module, rule-based agenda policy and template NLG. We conducted $1000$ dialogues and report the average number of turns across all dialogues. We focus on three measures: book rate, i.e., how often the system finalized a booking, success rate, i.e., the percentage of dialogues where all information requested by the user is provided by the system and bookings are successfully made, and lastly complete rate, i.e., the number of dialogues that are finished regardless of whether the booked entity matches the user criteria. We also report entity F1 and average number of turns across the simulated dialogues.

With the exception of AuGPT, the systems' dialogue policies require a dialogue state tracker (DST) for online interactions. For this purpose, we utilize a tracker with a joint goal accuracy of 52.26\% on the test set of MultiWOZ 2.1 \cite{van2020knowing}. This tracker is a recurrent neural model, which utilises attention and transformer based embeddings to extract important information from the dialogue. 
We perform lexicalization via handcrafted rules using the information from the dialogue state and database query. 
For handling incomplete lexicalizations due to empty database queries or a wrongly predicted domain by the policy, we replace the response with a generic ``I'm sorry, could you say that again?". This is equal to masking such actions
while neither punishing nor rewarding the policy.

\begin{table}[]
\centering
\small
\begin{tabular}{@{}clp{0.2\linewidth}p{0.2\linewidth}@{}}
\toprule
\multicolumn{1}{l}{}& & \multicolumn{1}{l}{SL} & \multicolumn{1}{l}{SL + PLAS} \\
\midrule
\multicolumn{1}{c}{\multirow{3}{*}{Corpus}} & Match & 66.06 & 83.94\\
\multicolumn{1}{c}{}& Success & 51.95 & 67.54\\
\multicolumn{1}{c}{}& BLEU& 0.17& 0.14 \\
\midrule
\multicolumn{1}{c}{\multirow{5}{*}{\begin{tabular}[c]{@{}c@{}}ConvLab\\ US\end{tabular}}} & Compl.& 37.42 & 47.02\\
\multicolumn{1}{c}{}& Success & 31.87 & 39.40\\
\multicolumn{1}{c}{}& Book& 19.12 & 36.74\\
\multicolumn{1}{c}{}& F1& 49.11 & 57.14\\
\multicolumn{1}{c}{}& Turns & 21.57 & 21.99\\
\bottomrule
\end{tabular}
\caption{Offline RL in latent space improves task-related metrics on both corpus and US evaluations. Results are averaged across 5 seeds.}
\label{tab:plas_lava}
\vspace{-10pt}
\end{table}

\begin{table*}[t]
\centering
\small
\begin{tabular}{@{}lcccr@{}}
\toprule
\multicolumn{1}{c}{\multirow{2}{*}{Policy}} & \multicolumn{3}{c}{Corpus Evaluation} & \multicolumn{1}{c}{\multirow{2}{*}{Critic Evaluation}} \\
\cmidrule{2-4}
\multicolumn{1}{c}{}& \multicolumn{1}{c}{Match} & \multicolumn{1}{c}{Success} & \multicolumn{1}{c}{BLEU} & \multicolumn{1}{c}{} \\
\midrule
MultiWOZ (Human)& \multicolumn{1}{r}{90.40 $\pm$ 1.82} & \multicolumn{1}{r}{82.30 $\pm$ 2.36} & \multicolumn{1}{c}{N/A} & 52.68 $\pm$ 0.02 \\
AuGPT & \multicolumn{1}{r}{83.30 $\pm$ 2.31} & \multicolumn{1}{r}{67.20 $\pm$ 2.91} & 0.17 & 52.45 $\pm$ 0.02  \\
LAVA+PLAS & \multicolumn{1}{r}{88.30 $\pm$ 1.99} & \multicolumn{1}{r}{73.40 $\pm$ 2.74} & 0.14 & 51.76 $\pm$ 0.03  \\
LAVA\_kl& \multicolumn{1}{r}{97.50 $\pm$ 1.14} & \multicolumn{1}{r}{94.80 $\pm$ 1.47} & 0.12 & 48.95 $\pm$ 0.08 \\
HDSA (gold)& \multicolumn{1}{r}{91.80 $\pm$ 1.70} & \multicolumn{1}{r}{82.50 $\pm$ 2.35} & 0.21 & 49.89 $\pm$ 0.08 \\
HDSA (pred)& \multicolumn{1}{r}{88.90 $\pm$ 1.95} & \multicolumn{1}{r}{74.50 $\pm$ 2.70} & 0.20 & 49.00 $\pm$ 0.09 \\
\bottomrule
\end{tabular}
\caption{Corpus-based evaluation metrics. 95\% confidence intervals are reported.}
\label{tab:corpus_eval}
\vspace{-5pt}
\end{table*}


\begin{table*}[t]
\centering
\small
\setlength{\tabcolsep}{4pt}
\begin{tabular}{@{}lcccccccc@{}}
\toprule
\multicolumn{1}{c}{\multirow{2}{*}{Policy}} & \multicolumn{5}{c}{ConvLab US Evaluation} & \phantom{} & \multicolumn{2}{c}{Human Evaluation} \\
\cmidrule{2-6} \cmidrule{8-9}
\multicolumn{1}{c}{} & \multicolumn{1}{c}{Compl.} & \multicolumn{1}{c}{Success} & \multicolumn{1}{c}{Book} & F1 & Avg. turn && \multicolumn{1}{c}{Success} & Rating \\
\midrule
AuGPT & \multicolumn{1}{l}{89.20 $\pm$ 1.92}& \multicolumn{1}{l}{83.30 $\pm$ 2.31} & \multicolumn{1}{l}{85.16 $\pm$ 3.34} & 81.03 $\pm$ 1.40 & 14.50 $\pm$ 0.41 && \multicolumn{1}{l}{90.75 $\pm$ 2.85} & 4.34 $\pm$ 0.08 \\
LAVA+PLAS & \multicolumn{1}{l}{54.20 $\pm$ 3.09}& \multicolumn{1}{l}{45.30 $\pm$ 3.09} & \multicolumn{1}{l}{61.18 $\pm$ 4.51} & 58.85 $\pm$ 2.25 & 23.54 $\pm$ 0.89 && \multicolumn{1}{l}{63.00 $\pm$ 4.75} & 3.34 $\pm$ 0.12 \\
LAVA\_kl & \multicolumn{1}{l}{49.20 $\pm$ 3.10}& \multicolumn{1}{l}{40.00 $\pm$ 3.04}& \multicolumn{1}{l}{63.20 $\pm$ 4.37} & 54.47 $\pm$ 2.24 & 26.64 $\pm$ 1.00 && \multicolumn{1}{l}{63.25 $\pm$ 4.74} & 3.44 $\pm$ 0.12 \\
HDSA (pred) & \multicolumn{1}{l}{36.70 $\pm$ 2.99}& \multicolumn{1}{l}{25.90 $\pm$ 2.71} & \multicolumn{1}{r}{6.67 $\pm$ 2.37}& 49.97 $\pm$ 2.23 & 31.32 $\pm$ 0.86 && \multicolumn{1}{l}{55.25 $\pm$ 4.89} & 3.09 $\pm$ 0.12\\
\bottomrule
\end{tabular}
\caption{Interactive evaluation metrics. 95\% confidence intervals are reported.}
\label{tab:online_eval}
\vspace{-10pt}
\end{table*}


\begin{table}[t]
\centering
\small
\setlength{\tabcolsep}{4pt}
\begin{tabular}{@{}lll|cc@{}}
\toprule
\multicolumn{3}{c}{\multirow{2}{*}{Fleiss' Kappa}}& \multicolumn{2}{c}{Human Evaluation}\\
\cmidrule{4-5}
\multicolumn{3}{c}{}& \multicolumn{1}{c}{Success} & \multicolumn{1}{c}{Rating} \\
\midrule
\multicolumn{1}{l|}{\multirow{4}{*}{Corpus-based}} & \multicolumn{1}{l|}{\multirow{3}{*}{Corpus}} & Match& -0.623 & -0.571 \\
\multicolumn{1}{l|}{}& \multicolumn{1}{l|}{}& Success& -0.460 & -0.397 \\
\multicolumn{1}{l|}{}& \multicolumn{1}{l|}{}& BLEU & 0.343 & 0.299\\ 
\cmidrule{2-5}
\multicolumn{1}{l|}{}& \multicolumn{2}{l|}{Critic} & \textbf{0.755} & \textbf{0.713}\\
\midrule
\multicolumn{1}{l|}{\multirow{5}{*}{Interactive}}& \multicolumn{1}{l|}{\multirow{5}{*}{US}} & Complete & 0.992 & 0.984\\
\multicolumn{1}{l|}{}& \multicolumn{1}{l|}{}& Success& 0.991 & 0.984 \\
\multicolumn{1}{l|}{}& \multicolumn{1}{l|}{}& Book & 0.789 & 0.802\\
\multicolumn{1}{l|}{}& \multicolumn{1}{l|}{}& F1 & 0.990 & 0.978\\
\multicolumn{1}{l|}{}& \multicolumn{1}{l|}{}& Turn & -0.967 & -0.956 \\
\bottomrule
\end{tabular}
\caption{Correlation between evaluation metrics and human judgements. Absolute values shows the strength of the correlation. Negative sign shows inverse correlation.}
\label{tab:correlation}
\vspace{-10pt}
\end{table}

\paragraph{Human evaluation}
Human evaluation is performed via DialCrowd~\cite{lee2018dialcrowd} connected to Amazon Mechanical Turk. The systems are set up identically as in the US evaluation, except that the systems are interacting with paid users instead of a US. Users are provided with a randomly generated user goal and are required to interact with our systems in natural language and to subsequently evaluate them. We ask the user whether their goal is fulfilled through the dialogue, indicating the success rate. We also ask them to rate the overall system performance on a Likert scale from 1 (worst) to 5 (best). For each system we collected 400 dialogues with human workers.

\section{Results and Analysis}


\subsection{Offline Critic for Optimization}

Table \ref{tab:plas_lava} shows the policy performance after shared multi-task SL training and the performance after subsequent offline RL training with PLAS, averaged over 5 seeds. We observe that offline RL in latent space with the critic estimate as reward signal improves task-related metrics on both corpus and US evaluation. The consistent improvement on offline and interactive evaluations is the result of critic's value estimate as reward signal, which we believe is noteworthy as the policy is never explicitly trained on either metric. 

Like policy gradient RL used by LAVA (Equation \ref{eq:reinforce}), PLAS leads to a decrease in BLEU score. This is quite common for end-to-end policies trained with RL following SL \cite{lubis2020lava}, however the decrease with PLAS is not as drastic. This signals that the policy retains more linguistic variety in the responses, since the reward signal does not overlook context mismatch and thus responses that are out of context are not rewarded. We include a dialogue example in Appendix \ref{sec:appendix:example} to demonstrate the context mismatch issue and how the offline critic addresses it.




\subsection{Offline Critic for Evaluation}

\paragraph{System performances across metrics}

Tables \ref{tab:corpus_eval} and \ref{tab:online_eval} present the corpus- and interaction-based evaluation results of LAVA+PLAS and our baselines. For completeness, we included the human policy, i.e., the behavior policy of the dataset, on the corpus-based evaluation. For LAVA+PLAS, we pick the best model out of the 5 seeds. For the baseline models, we utilize the released pre-trained parameters and re-run all evaluations.

The ranking of the systems differs depending on the evaluation metrics. With corpus-based success and match rates, LAVA far outperforms the other models and even human wizards.
This is expected, as LAVA\_kl is directly optimized with the corpus-based success rate as reward. In terms of BLEU, HDSA -- which is designed for generation with semantic action -- achieves the first rank. With critic evaluation, human policy achieves the highest score. 
The rankings for evaluation with user simulator and paid workers in Table \ref{tab:online_eval} are consistent, showing another trend entirely. AuGPT outperforms the other systems with a huge margin, LAVA+PLAS and LAVA\_kl show a narrower gap in performance compared to corpus-based metrics, while HDSA performs very poorly. The collected dialogues show that the language understanding and generation of AuGPT is superior to the other models, as it leverages a large pre-trained model as a base model and utilizes multiple dialogue corpora for fine-tuning. In other words, it is trained on orders of magnitude more data compared to the other systems. This results in a more natural interaction with both simulated and human users. 

It is interesting to note that the critic has a much narrower confidence interval compared to the other metrics. Although the values for some policies are seemingly close, the intervals show that the difference between most of the systems are statistically significant, except for LAVA\_kl and HDSA (gold).

\paragraph{Correlation with human judgements}

Table \ref{tab:correlation} lists pairwise correlation between human judgements and the automatic metrics. We differentiate between corpus-based metrics such as the standard match and success rates, BLEU and critic evaluation, with interactive metrics that require a form of user, either simulated or paid. 
Success rates of current standard evaluations have moderate inverse correlation with human judgements due to the context mismatch that occurs during its computation. On the other hand, the theoretically grounded value estimation by the offline critic has a strong correlation with human judgements, showing that our proposed method is a more suitable corpus-based metric to reflect the dialogue system performance. Our study confirms the weak correlation between BLEU and human ratings. All metrics computed based on interaction with US are strongly correlated with metrics from human evaluation. The number of turns is strongly but inversely correlated, which aligns with the intuition that the fewer turns the system needs to complete the dialogue, the better it is perceived by human users. This suggests that while existing US is far from fully imitating human behavior, it provides a good approximation to how the systems will perform when interacting with human users. We advocate that future works report on multiple evaluation metrics to provide a more complete picture of the dialogue system performance. 

Note that while US evaluation provides stronger correlations with human judgements, our proposed use of offline RL critic for evaluation has the benefit of being corpus- and model-independent, whereas for a new corpus and ontology, a new US would need to be designed and developed. Furthermore, an offline evaluation takes significantly less time to perform, making it an efficient choice for the iterative development process.

\subsection{Impact of Reward Signal on RL}

LAVA+PLAS and LAVA\_kl are the only two systems optimized via RL. We observe that they significantly outperform the other on the respective metric they received as reward signal during RL. However, when subjected to interactive evaluation, the gap between their performance is shrinking (see Table~\ref{tab:online_eval}). This shows on the one hand the power of reinforcement learning methods to optimize the given reward and on the other hand how important it is to define this reward correctly, warranting further research in both extrinsic and intrinsic reward modelling for dialogue~\cite{wesselmann,GeishauserHLLHF21}.

\section{Conclusion}

We propose the use of offline RL for dialogue evaluation based on static corpus. While offline RL critics are typically utilized for policy optimization, we show that they can be trained for any dialogue system as external evaluators that are corpus- and model-independent, while attaining strong correlation with human judgements, which we confirm via an interactive user trial. Not only does the offline RL critic provide a corpus-based metric that is reliable and efficient to compute, it also addresses a number of issues highlighted in the recently published NSF report~\cite{mehri2022report}. 
It is important to note that the proposed framework does not depend on the definition of states, action and rewards. So in principle, one could apply this method beyond task-oriented dialogue systems. For example, one could evaluate a number of chat-bots considering a corpus annotated only with level of engagement achieved in each dialogue and thus measure the level of engagement of the evaluated chat-bots.

\section*{Acknowledgements}
N. Lubis, C. van Niekerk, M. Heck and S. Feng are supported by funding provided by the Alexander von
Humboldt Foundation in the framework of the
Sofja Kovalevskaja Award endowed by the Federal Ministry of Education and Research. C.
Geishauser and H-C. Lin are supported
by funds from the European Research Council
(ERC) provided under the Horizon 2020 research
and innovation programme (Grant agreement No.
STG2018 804636). Google Cloud and HHU ZIM provided
computational infrastructure.

\bibliography{refs}

\begin{thebibliography}{42}
\expandafter\ifx\csname natexlab\endcsname\relax\def\natexlab#1{#1}\fi

\bibitem[{Budzianowski et~al.(2018)Budzianowski, Wen, Tseng, Casanueva, Stefan,
  Osman, and Ga{\v{s}}i\'c}]{budzianowski2018large}
Pawe{\l} Budzianowski, Tsung-Hsien Wen, Bo-Hsiang Tseng, I{\~n}igo Casanueva,
  Ultes Stefan, Ramadan Osman, and Milica Ga{\v{s}}i\'c. 2018.
\newblock Multi{WOZ} - a large-scale multi-domain {W}izard-of-{O}z dataset for
  task-oriented dialogue modelling.
\newblock In \emph{Proceedings of the 2018 Conference on Empirical Methods in
  Natural Language Processing (EMNLP)}.

\bibitem[{Byrne et~al.(2019)Byrne, Krishnamoorthi, Sankar, Neelakantan,
  Goodrich, Duckworth, Yavuz, Dubey, Kim, and Cedilnik}]{byrne2019taskmaster}
Bill Byrne, Karthik Krishnamoorthi, Chinnadhurai Sankar, Arvind Neelakantan,
  Ben Goodrich, Daniel Duckworth, Semih Yavuz, Amit Dubey, Kyu-Young Kim, and
  Andy Cedilnik. 2019.
\newblock Taskmaster-1: Toward a realistic and diverse dialog dataset.
\newblock In \emph{Proceedings of the 2019 Conference on Empirical Methods in
  Natural Language Processing and the 9th International Joint Conference on
  Natural Language Processing (EMNLP-IJCNLP)}, pages 4516--4525.

\bibitem[{Chen et~al.(2019)Chen, Chen, Qin, Yan, and
  Wang}]{chen2019semantically}
Wenhu Chen, Jianshu Chen, Pengda Qin, Xifeng Yan, and William~Yang Wang. 2019.
\newblock Semantically conditioned dialog response generation via hierarchical
  disentangled self-attention.
\newblock In \emph{Proceedings of the 57th Annual Meeting of the Association
  for Computational Linguistics}, pages 3696--3709.

\bibitem[{Dziri et~al.(2019)Dziri, Kamalloo, Mathewson, and
  Zaiane}]{dziri-etal-2019-evaluating}
Nouha Dziri, Ehsan Kamalloo, Kory Mathewson, and Osmar Zaiane. 2019.
\newblock \href {https://doi.org/10.18653/v1/N19-1381} {Evaluating coherence in
  dialogue systems using entailment}.
\newblock In \emph{Proceedings of the 2019 Conference of the North {A}merican
  Chapter of the Association for Computational Linguistics: Human Language
  Technologies, Volume 1 (Long and Short Papers)}, pages 3806--3812,
  Minneapolis, Minnesota. Association for Computational Linguistics.

\bibitem[{Eric et~al.(2019)Eric, Goel, Paul, Kumar, Sethi, Ku, Goyal, Agarwal,
  Gao, and Hakkani-Tur}]{eric2019multiwoz}
Mihail Eric, Rahul Goel, Shachi Paul, Adarsh Kumar, Abhishek Sethi, Peter Ku,
  Anuj~Kumar Goyal, Sanchit Agarwal, Shuyang Gao, and Dilek Hakkani-Tur. 2019.
\newblock Multi{WOZ} 2.1: A consolidated multi-domain dialogue dataset with
  state corrections and state tracking baselines.
\newblock \emph{arXiv preprint arXiv:1907.01669}.

\bibitem[{Fujimoto et~al.(2019)Fujimoto, Meger, and Precup}]{fujimoto2019off}
Scott Fujimoto, David Meger, and Doina Precup. 2019.
\newblock Off-policy deep reinforcement learning without exploration.
\newblock In \emph{International Conference on Machine Learning}, pages
  2052--2062. PMLR.

\bibitem[{Gabriel et~al.(2020)Gabriel, Liu, Gottardi, Eric, Khatri, Chadha,
  Chen, Hedayatnia, Rajan, Binici et~al.}]{gabriel2020further}
Raefer Gabriel, Yang Liu, Anna Gottardi, Mihail Eric, Anju Khatri, Anjali
  Chadha, Qinlang Chen, Behnam Hedayatnia, Pankaj Rajan, Ali Binici, et~al.
  2020.
\newblock Further advances in open domain dialog systems in the third {A}lexa
  prize socialbot grand challenge.
\newblock \emph{Alexa Prize Proceedings}.

\bibitem[{Geishauser et~al.(2021)Geishauser, Hu, Lin, Lubis, Heck, Feng, van
  Niekerk, and Gasic}]{GeishauserHLLHF21}
Christian Geishauser, Songbo Hu, Hsien{-}Chin Lin, Nurul Lubis, Michael Heck,
  Shutong Feng, Carel van Niekerk, and Milica Gasic. 2021.
\newblock \href {https://doi.org/10.1109/ASRU51503.2021.9687856} {What does the
  user want? information gain for hierarchical dialogue policy optimisation}.
\newblock In \emph{{IEEE} Automatic Speech Recognition and Understanding
  Workshop, {ASRU} 2021, Cartagena, Colombia, December 13-17, 2021}, pages
  969--976. {IEEE}.

\bibitem[{Gunasekara et~al.(2020)Gunasekara, Kim, D'Haro, Rastogi, Chen, Eric,
  Hedayatnia, Gopalakrishnan, Liu, Huang et~al.}]{gunasekara2020overview}
Chulaka Gunasekara, Seokhwan Kim, Luis~Fernando D'Haro, Abhinav Rastogi,
  Yun-Nung Chen, Mihail Eric, Behnam Hedayatnia, Karthik Gopalakrishnan, Yang
  Liu, Chao-Wei Huang, et~al. 2020.
\newblock Overview of the ninth dialog system technology challenge: Dstc9.
\newblock \emph{arXiv preprint arXiv:2011.06486}.

\bibitem[{Jaques et~al.(2019)Jaques, Ghandeharioun, Shen, Ferguson, Lapedriza,
  Jones, Gu, and Picard}]{jaques2019way}
Natasha Jaques, Asma Ghandeharioun, Judy~Hanwen Shen, Craig Ferguson, Agata
  Lapedriza, Noah Jones, Shixiang Gu, and Rosalind Picard. 2019.
\newblock Way off-policy batch deep reinforcement learning of implicit human
  preferences in dialog.
\newblock \emph{arXiv preprint arXiv:1907.00456}.

\bibitem[{Kulh{\'a}nek et~al.(2021)Kulh{\'a}nek, Hude{\v{c}}ek, Nekvinda, and
  Du{\v{s}}ek}]{kulhanek2021augpt}
Jon{\'a}{\v{s}} Kulh{\'a}nek, Vojt{\v{e}}ch Hude{\v{c}}ek, Tom{\'a}{\v{s}}
  Nekvinda, and Ond{\v{r}}ej Du{\v{s}}ek. 2021.
\newblock Au{GPT}: Dialogue with pre-trained language models and data
  augmentation.
\newblock \emph{arXiv preprint arXiv:2102.05126}.

\bibitem[{Kumar et~al.(2019)Kumar, Fu, Soh, Tucker, and
  Levine}]{kumar2019stabilizing}
Aviral Kumar, Justin Fu, Matthew Soh, George Tucker, and Sergey Levine. 2019.
\newblock Stabilizing off-policy q-learning via bootstrapping error reduction.
\newblock \emph{Advances in Neural Information Processing Systems}, 32.

\bibitem[{Kumar et~al.(2020)Kumar, Zhou, Tucker, and
  Levine}]{kumar2020conservative}
Aviral Kumar, Aurick Zhou, George Tucker, and Sergey Levine. 2020.
\newblock Conservative q-learning for offline reinforcement learning.
\newblock \emph{Advances in Neural Information Processing Systems},
  33:1179--1191.

\bibitem[{Lee et~al.(2018)Lee, Zhao, Black, and Eskenazi}]{lee2018dialcrowd}
Kyusong Lee, Tiancheng Zhao, Alan~W Black, and Maxine Eskenazi. 2018.
\newblock Dial{C}rowd: A toolkit for easy dialog system assessment.
\newblock In \emph{Proceedings of the 19th Annual SIGdial Meeting on Discourse
  and Dialogue}, pages 245--248.

\bibitem[{Lee and Esk{\'e}nazi(2012)}]{lees12}
Sungjin Lee and Maxine Esk{\'e}nazi. 2012.
\newblock {POMDP}-based let's go system for spoken dialog challenge.
\newblock In \emph{Proceedings of SLT}.

\bibitem[{Lillicrap et~al.(2016)Lillicrap, Hunt, Pritzel, Heess, Erez, Tassa,
  Silver, and Wierstra}]{lillicrap2016continuous}
Timothy~P Lillicrap, Jonathan~J Hunt, Alexander Pritzel, Nicolas Heess, Tom
  Erez, Yuval Tassa, David Silver, and Daan Wierstra. 2016.
\newblock Continuous control with deep reinforcement learning.
\newblock In \emph{ICLR (Poster)}.

\bibitem[{Lin et~al.(2021)Lin, Lubis, Hu, van Niekerk, Geishauser, Heck, Feng,
  and Gasic}]{lin-etal-2021-domain}
Hsien-chin Lin, Nurul Lubis, Songbo Hu, Carel van Niekerk, Christian
  Geishauser, Michael Heck, Shutong Feng, and Milica Gasic. 2021.
\newblock \href {https://aclanthology.org/2021.sigdial-1.47}
  {Domain-independent user simulation with transformers for task-oriented
  dialogue systems}.
\newblock In \emph{Proceedings of the 22nd Annual Meeting of the Special
  Interest Group on Discourse and Dialogue}, pages 445--456, Singapore and
  Online. Association for Computational Linguistics.

\bibitem[{Liu et~al.(2016)Liu, Lowe, Serban, Noseworthy, Charlin, and
  Pineau}]{liu2016not}
Chia-Wei Liu, Ryan Lowe, Iulian~Vlad Serban, Mike Noseworthy, Laurent Charlin,
  and Joelle Pineau. 2016.
\newblock How not to evaluate your dialogue system: An empirical study of
  unsupervised evaluation metrics for dialogue response generation.
\newblock In \emph{Proceedings of the 2016 Conference on Empirical Methods in
  Natural Language Processing}, pages 2122--2132.

\bibitem[{Lubis et~al.(2020)Lubis, Geishauser, Heck, Lin, Moresi, van Niekerk,
  and Ga{\v{s}}ic}]{lubis2020lava}
Nurul Lubis, Christian Geishauser, Michael Heck, Hsien-chin Lin, Marco Moresi,
  Carel van Niekerk, and Milica Ga{\v{s}}ic. 2020.
\newblock {LAVA}: Latent action spaces via variational auto-encoding for
  dialogue policy optimization.
\newblock In \emph{Proceedings of the 28th International Conference on
  Computational Linguistics}, pages 465--479.

\bibitem[{Mehri et~al.(2022)Mehri, Choi, D'Haro, Deriu, Eskenazi, Gasic,
  Georgila, Hakkani-Tur, Li, Rieser et~al.}]{mehri2022report}
Shikib Mehri, Jinho Choi, Luis~Fernando D'Haro, Jan Deriu, Maxine Eskenazi,
  Milica Gasic, Kallirroi Georgila, Dilek Hakkani-Tur, Zekang Li, Verena
  Rieser, et~al. 2022.
\newblock Report from the {NSF} future directions workshop on automatic
  evaluation of dialog: Research directions and challenges.
\newblock \emph{arXiv preprint arXiv:2203.10012}.

\bibitem[{Mehri and Eskenazi(2020{\natexlab{a}})}]{mehri2020unsupervised}
Shikib Mehri and Maxine Eskenazi. 2020{\natexlab{a}}.
\newblock Unsupervised evaluation of interactive dialog with dialo{GPT}.
\newblock In \emph{Proceedings of the 21th Annual Meeting of the Special
  Interest Group on Discourse and Dialogue}, pages 225--235.

\bibitem[{Mehri and Eskenazi(2020{\natexlab{b}})}]{mehri2020usr}
Shikib Mehri and Maxine Eskenazi. 2020{\natexlab{b}}.
\newblock {USR}: An unsupervised and reference free evaluation metric for
  dialog generation.
\newblock In \emph{Proceedings of the 58th Annual Meeting of the Association
  for Computational Linguistics}, pages 681--707.

\bibitem[{Mehri et~al.(2019)Mehri, Srinivasan, and
  Eskenazi}]{mehri2019structured}
Shikib Mehri, Tejas Srinivasan, and Maxine Eskenazi. 2019.
\newblock Structured fusion networks for dialog.
\newblock In \emph{Proceedings of the 20th Annual SIGdial Meeting on Discourse
  and Dialogue}, pages 165--177.

\bibitem[{Nekvinda and Du{\v{s}}ek(2021)}]{nekvinda2021shades}
Tom{\'a}{\v{s}} Nekvinda and Ond{\v{r}}ej Du{\v{s}}ek. 2021.
\newblock Shades of {BLEU}, flavours of success: The case of {M}ulti{WOZ}.
\newblock \emph{arXiv preprint arXiv:2106.05555}.

\bibitem[{Papineni et~al.(2002)Papineni, Roukos, Ward, and
  Zhu}]{papineni2002bleu}
Kishore Papineni, Salim Roukos, Todd Ward, and Wei-Jing Zhu. 2002.
\newblock {BLEU}: a method for automatic evaluation of machine translation.
\newblock In \emph{Proceedings of ACL}.

\bibitem[{Radford et~al.(2019)Radford, Wu, Child, Luan, Amodei, and
  Sutskever}]{radford2019language}
Alec Radford, Jeffrey Wu, Rewon Child, David Luan, Dario Amodei, and Ilya
  Sutskever. 2019.
\newblock Language models are unsupervised multitask learners.
\newblock \emph{OpenAI Blog}, 1(8):9.

\bibitem[{Ramachandran et~al.(2021)Ramachandran, Hashimoto, and
  Xiong}]{ramachandran2021causal}
Govardana~Sachithanandam Ramachandran, Kazuma Hashimoto, and Caiming Xiong.
  2021.
\newblock Causal-aware safe policy improvement for task-oriented dialogue.
\newblock \emph{arXiv preprint arXiv:2103.06370}.

\bibitem[{Rastogi et~al.(2020)Rastogi, Zang, Sunkara, Gupta, and
  Khaitan}]{rastogi2020towards}
Abhinav Rastogi, Xiaoxue Zang, Srinivas Sunkara, Raghav Gupta, and Pranav
  Khaitan. 2020.
\newblock Towards scalable multi-domain conversational agents: The
  schema-guided dialogue dataset.
\newblock In \emph{Proceedings of the AAAI Conference on Artificial
  Intelligence}, volume~34, pages 8689--8696.

\bibitem[{Schatzmann(2008)}]{schatzmann08a}
Jost Schatzmann. 2008.
\newblock \emph{{Statistical User and Error Modelling for Spoken Dialogue
  Systems}}.
\newblock Ph.D. thesis, University of Cambridge.

\bibitem[{Silver et~al.(2014)Silver, Lever, Heess, Degris, Wierstra, and
  Riedmiller}]{silver2014deterministic}
David Silver, Guy Lever, Nicolas Heess, Thomas Degris, Daan Wierstra, and
  Martin Riedmiller. 2014.
\newblock Deterministic policy gradient algorithms.
\newblock In \emph{International conference on machine learning}, pages
  387--395. PMLR.

\bibitem[{Stent et~al.(2005)Stent, Marge, and Singhai}]{stent2005evaluating}
Amanda Stent, Matthew Marge, and Mohit Singhai. 2005.
\newblock Evaluating evaluation methods for generation in the presence of
  variation.
\newblock In \emph{international conference on intelligent text processing and
  computational linguistics}, pages 341--351. Springer.

\bibitem[{Tseng et~al.(2021)Tseng, Dai, Kreyssig, and
  Byrne}]{tseng2021transferable}
Bo-Hsiang Tseng, Yinpei Dai, Florian Kreyssig, and Bill Byrne. 2021.
\newblock Transferable dialogue systems and user simulators.
\newblock In \emph{Proceedings of the 59th Annual Meeting of the Association
  for Computational Linguistics and the 11th International Joint Conference on
  Natural Language Processing (Volume 1: Long Papers)}, pages 152--166.

\bibitem[{Ultes et~al.(2017)Ultes, Budzianowski, Casanueva, Mrk\v{s}i\'{c},
  Rojas-Barahona, Su, Wen, Ga{\v s}i{\'{c}}, and Young}]{ultes2017domain}
Stefan Ultes, Pawe{\l} Budzianowski, Inigo Casanueva, Nikola Mrk\v{s}i\'{c},
  Lina~Maria Rojas-Barahona, Pei-Hao Su, Tsung-Hsien Wen, Milica Ga{\v
  s}i{\'{c}}, and Steve~J Young. 2017.
\newblock Domain-independent user satisfaction reward estimation for dialogue
  policy learning.
\newblock In \emph{INTERSPEECH}, pages 1721--1725.

\bibitem[{van Niekerk et~al.(2020)van Niekerk, Heck, Geishauser, Lin, Lubis,
  Moresi, and Gasic}]{van2020knowing}
Carel van Niekerk, Michael Heck, Christian Geishauser, Hsien-Chin Lin, Nurul
  Lubis, Marco Moresi, and Milica Gasic. 2020.
\newblock Knowing what you know: Calibrating dialogue belief state
  distributions via ensembles.
\newblock In \emph{Findings of the Association for Computational Linguistics:
  EMNLP 2020}, pages 3096--3102.

\bibitem[{Verma et~al.(2022)Verma, Fu, Yang, and Levine}]{verma2022chai}
Siddharth Verma, Justin Fu, Mengjiao Yang, and Sergey Levine. 2022.
\newblock Chai: A chatbot ai for task-oriented dialogue with offline
  reinforcement learning.
\newblock \emph{arXiv preprint arXiv:2204.08426}.

\bibitem[{Walker et~al.(1997)Walker, Litman, Kamm, and
  Abella}]{walker1997paradise}
Marilyn Walker, Diane Litman, Candace~A Kamm, and Alicia Abella. 1997.
\newblock {PARADISE}: A framework for evaluating spoken dialogue agents.
\newblock In \emph{35th Annual Meeting of the Association for Computational
  Linguistics and 8th Conference of the European Chapter of the Association for
  Computational Linguistics}, pages 271--280.

\bibitem[{{Wesselmann} et~al.(2019){Wesselmann}, {Wu}, and
  {Gašić}}]{wesselmann}
Paula {Wesselmann}, Yen-Chen {Wu}, and Milica {Gašić}. 2019.
\newblock \href {https://doi.org/10.1109/ICASSP.2019.8683033} {Curiosity-driven
  reinforcement learning for dialogue management}.
\newblock In \emph{ICASSP 2019 - 2019 IEEE International Conference on
  Acoustics, Speech and Signal Processing (ICASSP)}, pages 7210--7214.

\bibitem[{Wu et~al.(2019)Wu, Tucker, and Nachum}]{wu2019behavior}
Yifan Wu, George Tucker, and Ofir Nachum. 2019.
\newblock Behavior regularized offline reinforcement learning.
\newblock \emph{arXiv preprint arXiv:1911.11361}.

\bibitem[{Xie et~al.(2020)Xie, Harrison, and Finn}]{xie2020deep}
Annie Xie, James Harrison, and Chelsea Finn. 2020.
\newblock Deep reinforcement learning amidst lifelong non-stationarity.
\newblock \emph{arXiv preprint arXiv:2006.10701}.

\bibitem[{Zhao et~al.(2019)Zhao, Xie, and Eskenazi}]{zhao2019rethinking}
Tiancheng Zhao, Kaige Xie, and Maxine Eskenazi. 2019.
\newblock Rethinking action spaces for reinforcement learning in end-to-end
  dialog agents with latent variable models.
\newblock In \emph{Proceedings of the 2019 Conference of the North American
  Chapter of the Association for Computational Linguistics: Human Language
  Technologies, Volume 1 (Long and Short Papers)}, pages 1208--1218.

\bibitem[{Zhou et~al.(2020)Zhou, Bajracharya, and Held}]{PLAS_corl2020}
Wenxuan Zhou, Sujay Bajracharya, and David Held. 2020.
\newblock Plas: Latent action space for offline reinforcement learning.
\newblock In \emph{Conference on Robot Learning}.

\bibitem[{Zhu et~al.(2020)Zhu, Zhang, Fang, Li, Takanobu, Li, Peng, Gao, Zhu,
  and Huang}]{zhu2020convlab}
Qi~Zhu, Zheng Zhang, Yan Fang, Xiang Li, Ryuichi Takanobu, Jinchao Li, Baolin
  Peng, Jianfeng Gao, Xiaoyan Zhu, and Minlie Huang. 2020.
\newblock Conv{L}ab-2: An open-source toolkit for building, evaluating, and
  diagnosing dialogue systems.
\newblock In \emph{Proceedings of the 58th Annual Meeting of the Association
  for Computational Linguistics: System Demonstrations}, pages 142--149.

\end{thebibliography}
\bibliographystyle{acl_natbib}

\appendix

\section{Example Dialogue}
\label{sec:appendix:example}
Table \ref{tab:example} shows an example dialogue accompanied by the corresponding offline critic Q estimation and corpus-based success rate.

\begin{table*}[]
\centering
\small
\setlength{\tabcolsep}{4pt}
\begin{tabular}{|l|p{0.7\linewidth}|l|l|}
\hline
 & Utterances& Q& Corpus succ. \\ \hline
User & can you help me find a {[}value\_pricerange{]} place to stay in the {[}value\_area{]} part of town?&&\\ \cline{1-2}
True & sure. there are {[}value\_count{]} guesthouses there. i would be happy to book {[}value\_count{]} for you if you like. &&\\
Pred & the {[}hotel\_name{]} is located at {[}hotel\_address{]} , postcode {[}hotel\_postcode{]}. the phone number is {[}hotel\_phone{]}.& 0.4730 &\\ \cline{1-3}
User & does it have a star rating of {[}value\_count{]}?&&\\ \cline{1-2}
True & no. they all have {[}value\_count{]} stars. &&\\
Pred & \textit{the {[}hotel\_name{]} is a guesthouse in the {[}value\_area{]} area. the phone number is {[}hotel\_phone{]} and the postcode is {[}hotel\_postcode{]}.}& 0.4569 &\\ \cline{1-3}
User & then find me {[}value\_count{]} in the {[}value\_pricerange{]} price range.&&\\ \cline{1-2}
True & all of the accommodations in the {[}value\_area{]} are in the {[}value\_pricerange{]} price range. would you like to try a different part of town?&&\\
Pred & \textit{the {[}hotel\_name{]} is located at {[}hotel\_address{]} , postcode {[}hotel\_postcode{]}. the phone number is {[}hotel\_phone{]}.}& 0.4478 &\\ \cline{1-3}
User & find me a nice {[}value\_count{]} and book for {[}value\_count{]} people and {[}value\_count{]} nights from {[}value\_day{]}&&\\  \cline{1-2}
True & actually , the {[}hotel\_name{]} is a {[}value\_count{]} start hotel in the {[}value\_area{]} area. it s in the {[}value\_pricerange{]} price range. how does that sound?&&\\
Pred & \textit{i have booked your room at the {[}hotel\_name{]} for {[}value\_count{]} nights starting {[}value\_day{]}. your reference number is {[}hotel\_reference{]}. is there anything else i can help you with?} & 0.4461 &\\\cline{1-3}
User & that sounds perfect. can you reserve that and provide me with the reference number? &&\\ \cline{1-2}
True & okay , you are all set. the hotel is booked and the reference number is {[}hotel\_reference{]}. can i help you with anything else today? &&\\
Pred & the reference number is {[}hotel\_reference{]}.& 0.4205 &\\ \cline{1-3}
User & thank you. i am also looking for a restaurant that serves {[}value\_food{]} and that is in the same area as the hotel&&\\ \cline{1-2}
True & {[}restaurant\_name{]} is an {[}value\_pricerange{]} restaurant in the same area as the hotel , would you like me to make reservations for you?&&\\ 
Pred & {[}restaurant\_name{]} is located at {[}restaurant\_address{]} , {[}restaurant\_postcode{]}. the phone number is {[}restaurant\_phone{]}. & 0.4167 &\\ \cline{1-3}
User & yes. same group of people , at {[}value\_time{]} on the same day please.&&\\ \cline{1-2}
True & booking was successful. the table will be reserved for {[}value\_count{]} minutes. reference number is {[}restaurant\_reference{]}. what else can i help you with today?&&\\
Pred & i have booked you a table at {[}restaurant\_name{]}. the reference number is {[}restaurant\_reference{]}. can i help you with anything else? & 0.4095 &\\ \cline{1-3}
User & that was all i needed. thanks. bye.&&\\ \cline{1-2}
True & ok! have a great day! &&\\
Pred & you are welcome. have a great day!& 0.4161 & 1.0 \\\hline
\end{tabular}
\caption{Example dialogue comparing offline critic estimate and corpus-based success. "True" denotes responses taken from the corpus, and "Pred" responses from the policy, in this case we use LAVA\_kl with which context mismatch often occurs. Note that Q prediction takes "User" and "True" utterances from the beginning up to the previous turn, and "User" and "Pred" of current turn. On the other hand, Corpus-based success takes on "User" and "Pred" utterances for all turns. Predicted responses in italic highlight the context mismatch that can occur when pseudo-dialogue is constructed for dialogue success computation. This is however ignored and the dialogue is considered successful, since all necessary requestable slots are generated by the system. On the other hand, the Q-estimate shows a decrease in value, and the policy is given a lower reward signal for the same dialogue.}
\label{tab:example}
\end{table*}
\end{document}